\documentclass[times,twocolumn,final,authoryear]{elsarticle}

\usepackage{prletters}
\usepackage{framed,multirow}

\usepackage{amssymb}
\usepackage{latexsym}
\usepackage{amsmath}
\usepackage{algpseudocode}

\usepackage{natbib}

\usepackage{url}
\usepackage{xcolor}
\definecolor{newcolor}{rgb}{.8,.349,.1}

\usepackage{fancyhdr}
\begin{document}

\thispagestyle{empty}

\clearpage
\thispagestyle{empty}
\ifpreprint
  \vspace*{-1pc}
\fi

\begin{frontmatter}

\title{Enhanced Variational Inference with Dyadic Transformation}

\author[1,2]{Sarin \snm{Chandy}} 

\author[1]{Amin \snm{Rasekh}\corref{cor1}}
\cortext[cor1]{Corresponding author: 
  Tel.: +1-919-845-4000;  
  fax: +1-919-845-4000;}
\ead{amin.rasekh@xyleminc.com}

\address[1]{Xylem Inc., 817 West Peachtree Street, Atlanta, GA 30308, USA}
\address[2]{The University of Chicago, 5807 S Woodlawn Ave, Chicago, IL 60637, USA}

\begin{abstract}
Variational autoencoder is a powerful deep generative model with variational inference. The practice of modeling latent variables in the VAE's original formulation as normal distributions with a diagonal covariance matrix limits the flexibility to match the true posterior distribution. We propose a new transformation, dyadic transformation (DT), that can model a multivariate normal distribution. DT is a single-stage transformation with low computational requirements. We demonstrate empirically on MNIST dataset that DT enhances the posterior flexibility and attains competitive results compared to other VAE enhancements. \\
\\

\textbf{Keywords:} Autoendcoder; Generative models; Variational inference; Dyadic transformation. \\

\\

source code available at  \url{https://github.com/sarin1991/DyadicFlow}

\end{abstract}

\end{frontmatter}

%\linenumbers

%% main text
\section{Introduction}
A VAE is a deep generative model with variational inference. A generative model is an unsupervised learning approach that is able to learn a domain by processing a large amount of data from it and then generate new data like it \citep{hinton1997generative, yu2018deep}. VAE, together with Generative Adversarial Networks \citep{goodfellow2016deep} and Deep Autoregressive Networks \citep{gregor2013deep}, are amongst the most powerful and popular generative model techniques. VAE has been successfully applied in many domains, such as image processing \citep{pu2016variational}, natural language processing \citep{semeniuta2017hybrid}, and cybersecurity \citep{chandy2018cyberattack}.

A VAE works by maximizing a variational lower bound of the likelihood of the data \citep{kingma2013auto}. A VAE has two halves: a recognition model (an encoder) and a generative model (a decoder). The recognition model learns a latent representation of the input data, and the generative model learns to transform this representation back into the original data. The recognition and generative models are jointly trained by optimizing the probability of the input data using stochastic gradient ascent. 

Application of the VAE involves selection of an approximate posterior distribution for the latent variables. This decision determines the flexibility and tractability of the VAE, and hence the quality and efficiency of the inference made, and poses a core challenge in variational inference. Conventionally, the choice is the normal distribution with a diagonal covariance matrix. This pick helps with computation efficiency but limits the flexibility to match the true posterior. We introduce a new transformation, DT, which approximates the posterior as a normal distribution with full covariance. DT offers theoretical advantages of model flexibility, parallelizability, scalability, and efficiency, which together provide a clear improvement in VAE for its wider adoption for statistical inference in the presence of large, complex datasets.

\section{Variational Autoencoder}
\subsection{Formulation}
Let \textbf{x} be a (set of) observed variables, \textbf{z} a (set of) continuous, stochastic latent variables that represent their encoding, and $p(\textbf{x},\textbf{z})$ the parametric model of their joint distribution. The observations of \textbf{x} (datapoints) are generated by a random process, which involves the unobserved random variables \textbf{z}. The encoder network with parameters $\phi$ encodes the given dataset with an approximate posterior distribution given by $q_\phi(\textbf{z}|\textbf{x})$ defined over the latent variables, while the decoder network with parameters $\theta$ decodes \textbf{z} into \textbf{x} with probability $p_\theta(\textbf{x}|\textbf{z})$. The encoder tries to approximate the true but intractable posterior represented as $p_\theta(\textbf{z}|\textbf{x})$. By assuming a standard normal prior for the decoder and given a dataset $\textbf{X}$, we can optimize the network parameters by maximizing the log-probability of the data $p_\theta(\textbf{X})$, i.e., to maximize
\begin{equation}
\log p(\textbf{X})=\log p \big(\textbf{x}^{(1)}, ..., \textbf{x}^{(N)}\big)=\sum_{i=1}^{N}\log p\big(\textbf{x}^{(i)}\big) 
\end{equation}

where, given our approximation to the true posterior distribution, for each datapoint \textbf{x} we can write

\begin{equation}
\log p_\theta(\textbf{x})\geq \mathbb{E}_{q_\phi(\textbf{z}|\textbf{x})}\Big[\log p_\theta(\textbf{x}|\textbf{z}) \Big] - D_{KL}\Big(  q_\phi(\textbf{z}|\textbf{x}) || p_\theta (\textbf{z}) \Big)
\end{equation}

The RHS term is denoted as $\mathcal{L} \big( \theta,\phi;\textbf{x} \big)$. Because KL divergence $D_{KL}(.)$ is always non-negative, it can be written as follows and is the (variational) lower bound on the marginal likelihood of datapoint \textbf{x}

\begin{equation}
\mathcal{L} \big( \theta,\phi;\textbf{x} \big)=\log p_\theta(\textbf{x}) - D_{KL}\Big(  q_\phi(\textbf{z}|\textbf{x}) || p_\theta (\textbf{z}|\textbf{x}) \Big)
\end{equation}

Therefore, maximizing the lower bound will simultaneously increase the probability of the data and reduce divergence from the true posterior. Thus, we would like to maximize it w.r.t. the encoder and decoder parameters, $\theta$ and $\phi$, respectively.

\subsection{Need for model flexibility}
The encoder and decoder in a VAE are conventionally modeled using the normal distribution with a diagonal covariance matrix, i.e., $\mathcal{N}(\boldsymbol{\mu} ,\mathrm{diag}(\boldsymbol{\sigma}^{2}))$, where $\boldsymbol{\mu}$ and $\boldsymbol{\sigma}$ are commonly nonlinear functions parametrized by neural networks. This practice is mainly driven by the requirements for computational tractability. It, however, limits flexibility of the model, especially in the case of the encoder where the encoder will not be able to learn the true posterior distribution. 

\section{Dyadic Transformation}
\subsection{Motivation}
Theoretically, the approximate model will be significantly more flexible if it is modeled as a multivariate normal distribution with a full covariance matrix.

A linear transformation matrix \textbf{B} of size $n \times n$ applied on an $n$-dimensional normal distribution $\textbf{Y} \sim \mathcal{N}(\boldsymbol{\mu},\boldsymbol{\sigma}^{2})$ produces another normal distribution $\textbf{G} \sim \mathcal{N}(\textbf{B}\boldsymbol{\mu},\,\textbf{B}\boldsymbol{\sigma}^{2}\textbf{B}^T)$. Thus, although \textbf{Y} is a normal distribution with diagonal covariance, its transformation through \textbf{B} would result in a multivariate normal distribution:

\begin{equation}
\textbf{G} = \textbf{B} \textbf{Y}
\end{equation}

This transformation matrix \textbf{B} introduces $O(n^2)$ number of new parameters. In order to utilize this transformation in our generative model, we would need to compute the log-probability and KL divergence of \textbf{G}. These computations do not scale well with the size of \textbf{B}. 

To overcome this issue, we define the transformation matrix \textbf{B} as follows:

\begin{equation}
\textbf{B}=\textbf{I}+ \epsilon \textbf{UV}
\end{equation}

where \textbf{I} is an identity matrix, $\epsilon$ is a scalar parameter, \textbf{U} is an $n \times k$ matrix, and \textbf{V} is a $k \times n$ matrix. Here $k$ is a model hyper-parameter that can be adjusted to set the trade-off between algorithm flexibility and computational efficiency.

In what follows, we show that this affine transformation gives the higher flexibility desired without introducing much additional computational complexity and thus it scales well with \textit{n}.

\begin{figure}[h]
  \centering
    \includegraphics[width=0.5\textwidth]{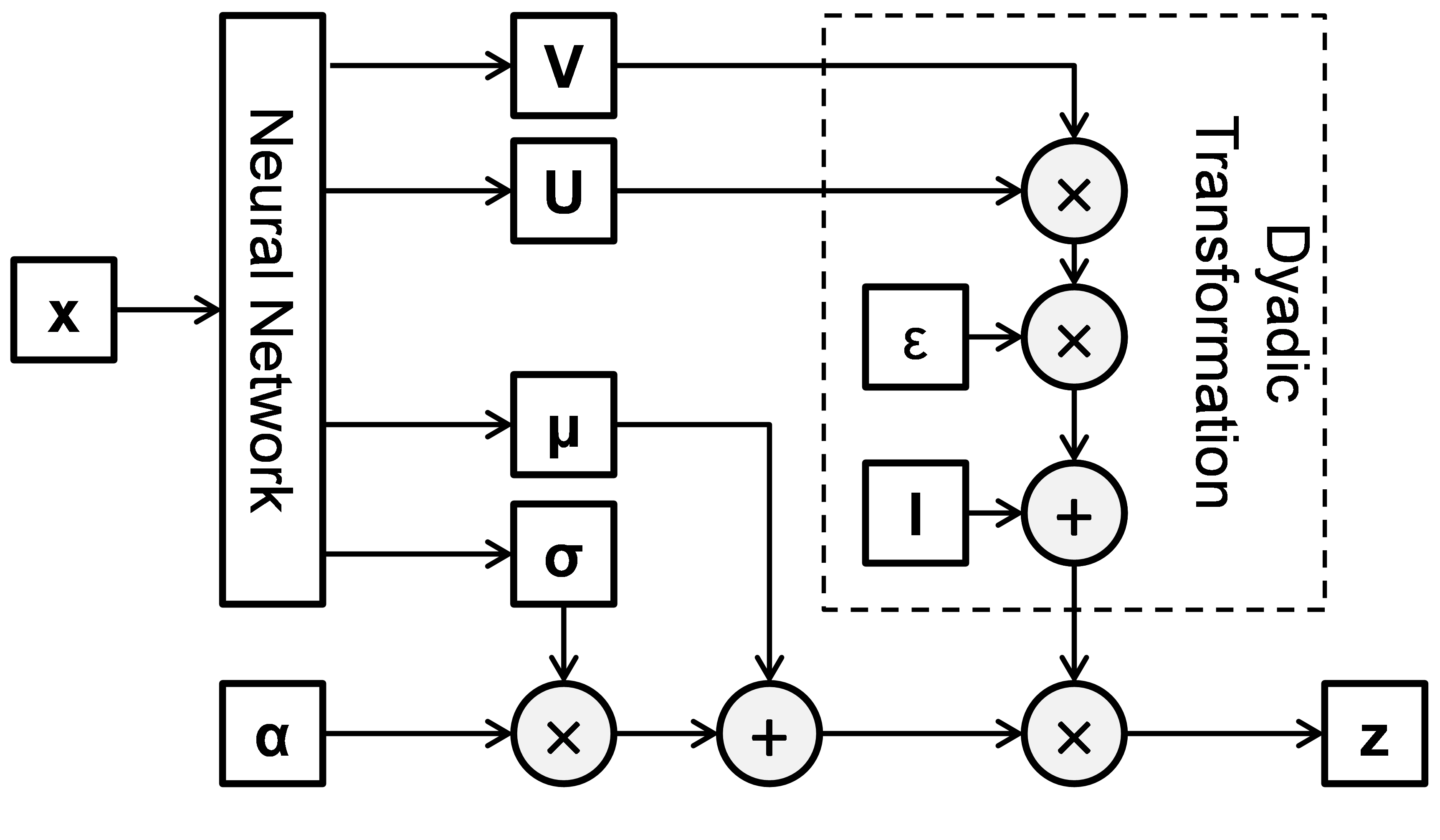}
  \caption[width=3cm, height=4cm]{A sample \textbf{z} drawn from a VAE model with dyadic transformation for an input \textbf{x} involves a simple sequence of matrix addition and multiplication operations where $\boldsymbol{\alpha} \sim \mathcal{N} (\boldsymbol{0},\boldsymbol{1})$.}
\end{figure}

\subsection{Efficient calculation of matrix determinant and inverse}
Computing the log-probability and KL Divergence of the generative model involves the calculation of the determinant and inverse of the dyadic transformation matrix. We show that these operations can be efficiently computed with the help of the following theorems: \\

\textbf{Theorem 1.} (Sherman-Morrison-Woodbury). Given four matrices \textbf{A}, \textbf{U}, \textbf{C}, and \textbf{V}, 

\begin{equation}
(\textbf{A}+\textbf{UCV})^{-1}=\textbf{A}^{-1}-\textbf{A}^{-1}\textbf{U}(\textbf{C}^{-1}+\textbf{VA}^{-1}\textbf{U})^{-1}\textbf{VA}^{-1}
\end{equation}

if the matrices are of conformable sizes and also if the matrices \textbf{A} and  $\textbf{C}^{-1}+\textbf{VA}^{-1}\textbf{U}$ are invertible \citep{woodbury1950inverting}. With the help of this theorem, we can efficiently calculate the inverse for the Dyadic Transformation matrix \textbf{B}. \\

\textbf{Theorem 2.} (Sylvester's Determinant Identity). Given two matrices \textbf{U} and \textbf{V} of sizes $m \times n$ and $n \times m$, 

\begin{equation}
\det(\textbf{I}_m + \textbf{UV})=\det(\textbf{I}_n+\textbf{VU}),
\end{equation}

where $\mathbf{I}_m$ and $\mathbf{I}_n$ are identity matrices of orders $m$ and $n$, respectively \citep{sylvester1851xxxvii}. This theorem relates the determinant of an $n \times n$ matrix with the determinant of an $m \times m$ matrix, which is very useful in regimes were $n\gg m$. We use this property to make the determinant calculations of \textbf{B} computationally tractable.

\subsection{KL divergence between two normal distributions}
Using the above theorems we show that the KL divergence for a multivariate normal distribution obtained using Dyadic Transformation can be efficiently computed. 

KL divergence between the independent normal posterior and standard normal prior can be written as \citep{kingma2013auto}

\begin{equation}
D_{KL}\big( q(\textbf{z}|\textbf{x})||p(\textbf{z}) \big)=\frac{1}{2} \sum_{j=1}^{J} \big( 1 + \log p\big(\boldsymbol{\sigma}_j^2\big) - \boldsymbol{\mu} _j^2 - \boldsymbol{\sigma}_j^2 \big)
\end{equation}

where $J$ is the dimensionality of \textbf{z}. 
We can show that in general the KL divergence between two normal distributions, with means $\boldsymbol{\mu}_0$ and $\boldsymbol{\mu}_1$, and covariance matrices $\boldsymbol{\Sigma}_0$ and $\boldsymbol{\Sigma}_1$, is \citep{duchi2007derivations}:

  \begin{equation}
    \begin{aligned}[b]
        & D_{KL}(\mathcal{N}_0||\mathcal{N}_1)= \frac{1}{2} \times \\
        &  \big(  \mathrm{Tr}(\boldsymbol{\Sigma}_1^{-1}\boldsymbol{\Sigma}_0) + (\boldsymbol{\mu}_1-\boldsymbol{\mu}_0)^T \boldsymbol{\Sigma} _1^{-1} (\boldsymbol{\mu}_1-\boldsymbol{\mu}_0)-J + \ln(\det(\boldsymbol{\Sigma}_1/\boldsymbol{\Sigma}_0 ) \big)
    \end{aligned}
  \end{equation}

Given that in our case, $q(\textbf{z}|\textbf{x}) \sim \mathcal{N}(\boldsymbol{\mu} ,\boldsymbol{\Sigma})$ and $p(\textbf{z}) \sim \textbf{}(0,1)$, we can write

\begin{equation}
D_{KL} \big( q(\textbf{z}|\textbf{x}) || p(\textbf{z}) \big)=\frac{1}{2} \big( \mathrm{Tr}(\boldsymbol{\Sigma}) + \boldsymbol{\mu}^T\boldsymbol{\mu}-J - \ln(\det(\boldsymbol{\Sigma})) \big)
\end{equation}

We observe that calculation of KL divergence also involves the calculation of $\det(\boldsymbol{\Sigma})$. This is performed efficiently using the Sylvester's determinant theorem.

\subsection{Calculation of the gradient of matrix determinant and inverse}
Given a matrix \textbf{D}, the derivative of the inverse and determinant of \textbf{D} w.r.t. a variable \textbf{t} can be calculated as

\begin{equation}
\frac{\partial \textbf{(D)}^{-1}}{\partial t}=-\textbf{D}^{-1}\frac{\partial \textbf{D}}{\partial t}\textbf{D}^{-1}
\end{equation}

\begin{equation}
\frac{\partial \det (\textbf{D})}{\partial t}=\det(\textbf{D})\mathrm{Tr}(\textbf{D}^{-1} \frac{\partial \textbf{D}}{\partial t})
\end{equation}

We make a key observation from the two derivative equations above. That is, given the determinant and inverse of a matrix are finite, their gradients will also be finite. Calculation of either derivatives thus may not lead to numerical instability even if the matrix is initialized randomly.

Also from Equations (11) and (12) we can show that for the Dyadic Transformation matrix $\textbf{B}$

\begin{equation}
\det (\textbf{B})= 1 + \epsilon \mathrm{Tr}(\textbf{UV})+O(\epsilon^2)
\end{equation}

\begin{equation}
\textbf{B}^{-1}=\textbf{I}- \epsilon \textbf{UV}+O(\epsilon^2)
\end{equation}

An important observation here is that if the value of $\epsilon$ is small enough then the determinant and inverse of the dyadic transformation matrix will be finite. This observation was crucial for us in order to make the make the numerical computations stable. \\

\textbf{Pseudo code for VAE with Dyadic Transformation}
\begin{algorithmic}[1]

\Repeat
    \textbf{X}$^M \leftarrow$ Random minibatch of $M$ datapoints
    
    $\boldsymbol{\alpha} \leftarrow$ Random samples from noise distribution 
    
    $\textbf{U} , \textbf{V}, \boldsymbol{\mu}, \boldsymbol{\sigma} \leftarrow$ Encoder NN $(\textbf{X},\theta)$
    
    $\textbf{Y} \leftarrow \boldsymbol{\mu} + \boldsymbol{\alpha} \times \boldsymbol{\sigma}$
    
    $\textbf{z} \leftarrow (\textbf{I}+ \epsilon \textbf{\textbf{UV}}) \textbf{Y}$
    
    g $\leftarrow \nabla _{\theta,\phi}\hat{\mathcal{L}}^M (\theta,\phi;\textbf{X}^M, \textbf{z})$
    
    $\theta , \phi \leftarrow$ Update parameters using gradient g
\Until{convergence of parameters ($\theta$, $\phi$)}

\State \textbf{return} $\theta$, $\phi$

\end{algorithmic}

\section{Related Work}
Many recent strategies proposed to improve flexibility of inference models are based on the concept of normalizing flows, introduced by \citep{rezende2015variational} in the context of stochastic variational inference. Members of this family build a flexible variational posterior by starting with a conventional normal distribution for generating the latent variables and then applying a chain of invertible transformations, such as Householder transformation \citep{tomczak2016improving} and inverse autoregressive transformation \citep{kingma2016improving}. Our proposed strategy requires only a single transformation and can be applied to both the encoder and the decoder.

\section{Experiments}
We conducted experiments on MNIST dataset to empirically evaluate our approach. MNIST is a dataset of 60,000 training and 10,000 test images of handwritten digits with a resolution of 28 $\times$ 28 pixels \citep{lecun1998gradient}. The dataset was dynamically binarized as in \citep{salakhutdinov2008quantitative}.

Our model had 50 stochastic units each and the encoder and decoder were parameterized by a two-layer feed forward network with 500 units each. The model was trained using ADAM gradient-based optimization algorithm \citep{kingma2015adam} with a mini-batch size of 128. For the Dyadic Transformation matrix B we used a value of 0.001 for ${\epsilon}$.

The results of the experiments are presented in Table 1. The results indicate that our proposed strategy is able to obtain competitively low log-likelihoods despite its inherent simplicity and low computational requirements. Compared to VAE, DT adds an additional computational cost of $O(k^{2.37})$ which is primarily for the determinant calculation. Hence, for smaller values of \textit{k}, DT does not add any computational cost. Also the memory requirements for DT is $O(kn)$ which is also reasonable for small values of \textit{k}.  

Our idea is fundamentally different from the other strategies for improving VAE since it does not belong to the existing large family of normalizing flow transformations. Thus, it holds promise for creating a new family of strategies for building flexible distributions in the context of stochastic variational inference.

\begin{table} [h]
  \caption{Lower bound of the marginal log-likelihood for MNIST test dataset for the regular VAE and VAE with our dyadic transformation, HF \citep{tomczak2016improving}, NF \citep{rezende2015variational}, and HVI \citep{salimans2015markov}. Listed are averages across 5 optimization runs. $T$ denotes the length of the flows. $T$ does not apply to Dyadic Transformation method because it is a single-step transformation.}
  \label{MNIST-results}
  \centering
  \begin{tabular}{lll}

    Model     & $\leq \log p(\textbf{x})$  \\

    VAE & -89.93 \\
    VAE+DT (k=10) & -88.24 \\
    VAE+DT (k=20) & -88.00 \\
    VAE+DT (k=50) & \textbf{-87.42} \\

    VAE+NF (\textit{T}=80) & -85.1 \\
    VAE+HF (\textit{T}=10) & -87.68 \\
    VAE+HVI (\textit{T}=8) & -88.30 \\
  \end{tabular}
\end{table}

\section{Conclusion}
We presented Dyadic Transformation, a new transformation that builds flexible multivariate distribution to enhance variational inference without sacrificing computational tractability. Our elegantly-simple idea boosts model flexibility with only a single transformation step. The empirical experiments obtained indicated objectively that DT increases VAE performance and its results are competitive compared to the family of normalizing flows, which involve multiple levels of transformation. Our transformation can be readily integrated with the methods in this family to collectively build powerful hybrids. Dyadic Transformation can also be straightforwardly applied to the decoder to obtain even more significant performance gains. It can also be applied to binary data by modifying a Restricted Boltzmann Machine. These theoretical advantages will be explored in future research.

%\bibliographystyle{model2-names}
%\bibliography{refs}

%This is where your bibliography is generated. Make sure that your .bib file is actually called library.bib
\bibliographystyle{plainnat}
\bibliography{dfvae}

%This defines the bibliographies style. Search online for a list of available styles.

\end{document}